\documentclass[11pt]{article}

\makeatletter
\def\input@path{{AIPaper_arxiv/}{}}
\makeatother

\usepackage[utf8]{inputenc}
\usepackage[T1]{fontenc}
\usepackage[letterpaper,margin=1in]{geometry}
\usepackage{microtype}
\usepackage{natbib}
\usepackage{amsmath}
\usepackage{amssymb}
\usepackage{amsthm}
\usepackage{mathtools}
\usepackage{bm}
\usepackage{graphicx}
\graphicspath{{AIPaper_arxiv/}{}}
\usepackage{booktabs}
\usepackage{array}
\usepackage{multirow}
\usepackage{float}
\usepackage{xcolor}
\usepackage{colortbl}
\usepackage{tcolorbox}
\tcbuselibrary{breakable}
\usepackage{enumitem}
\usepackage{algorithm}
\usepackage{algpseudocode}
\usepackage{listings}
\usepackage[colorlinks=true,linkcolor=blue,citecolor=brown,urlcolor=blue]{hyperref}
\usepackage{url}

\makeatletter
\def\theHALG@line{\thealgorithm.\arabic{ALG@line}}
\makeatother

\lstdefinestyle{reprocode}{
  basicstyle=\ttfamily\tiny,
  columns=fullflexible,
  keepspaces=true,
  showstringspaces=false,
  breaklines=true,
  breakatwhitespace=false,
  tabsize=4,
  frame=single,
  rulecolor=\color{black!25},
  xleftmargin=0.5em,
  xrightmargin=0.5em
}

\newtcolorbox{examplebox}[1]{
  breakable,
  colback=gray!5,
  colframe=gray!60,
  fonttitle=\bfseries\small,
  title=#1,
  boxrule=0.5pt,
  arc=2pt,
  left=6pt, right=6pt, top=4pt, bottom=4pt
}

\newtcolorbox{resultplaceholder}[1]{
  colback=gray!4,
  colframe=gray!55,
  fonttitle=\bfseries\small,
  title=#1,
  boxrule=0.5pt,
  arc=2pt,
  left=6pt, right=6pt, top=5pt, bottom=5pt
}

\newif\ifhavecleveref
\IfFileExists{cleveref.sty}{\haveclevereftrue}{\haveclevereffalse}
\ifhavecleveref
  \usepackage{cleveref}
\else
  \makeatletter
  \@namedef{crefw@sec}{Section}%
  \@namedef{crefw@fig}{Figure}%
  \@namedef{crefw@tab}{Table}%
  \@namedef{crefw@alg}{Algorithm}%
  \@namedef{crefw@eq}{Equation}%
  \@namedef{crefw@prop}{Proposition}%
  \@namedef{crefw@lem}{Lemma}%
  \@namedef{crefw@cor}{Corollary}%
  \@namedef{crefw@def}{Definition}%
  \@namedef{crefw@app}{Appendix}%
  \def\cref@split#1:#2\@nil{#1}%
  \newcommand{\@dolabel}[1]{%
    \edef\@thepfx{\cref@split#1\@nil}%
    \@ifundefined{crefw@\@thepfx}{}{\csname crefw@\@thepfx\endcsname~}\ref{#1}}%
  \newcommand{\Cref}[1]{\@dolabel{#1}}%
  \newcommand{\cref}[1]{\@dolabel{#1}}%
  \makeatother
\fi

\theoremstyle{plain}

\theoremstyle{definition}

\newcommand{\Z}{\mathbb{Z}}
\newcommand{\R}{\mathbb{R}}
\newcommand{\conv}{\operatorname{conv}}
\newcommand{\nrp}{n_{\mathrm{rp}}}

\newcommand{\SearchLocalStarts}{1{,}000}
\newcommand{\SearchLocalThreeImproved}{990}
\newcommand{\SearchLocalFourImproved}{991}

\newcommand{\SearchLocalThreeScoreOne}{27}
\newcommand{\SearchLocalFourScoreOne}{25}

\newcommand{\SearchExhaustiveSeeds}{50}
\newcommand{\SearchExhaustiveThreeTested}{159{,}500}
\newcommand{\SearchExhaustiveFourTested}{499{,}000}
\newcommand{\SearchExhaustiveThreeConvex}{85{,}913}
\newcommand{\SearchExhaustiveFourConvex}{173{,}355}

\newcommand{\SearchOEMutations}{20}
\newcommand{\SearchOECandidatesPerMutation}{32}
\newcommand{\SearchOECandidates}{640}
\newcommand{\SearchOECalls}{1{,}000}
\newcommand{\SearchOEThreeFirstBest}{7}
\newcommand{\SearchOEFourFirstBest}{11}

\title{RadonBench: A Benchmark of Classical and Learning-Guided Search Techniques for Problems in Combinatorial Geometry}

\author{%
  Xiaoan Xu\thanks{Department of Electrical and Computer Engineering, Duke University.},
  Jesus A. De Loera\thanks{Department of Mathematics, University of California, Davis.},
  Ethan X. Fang\thanks{Department of Biostatistics and Bioinformatics, Duke University.},
  Junwei Lu\thanks{Department of Biostatistics, Harvard T.H. Chan School of Public Health.},
  Luze Xu\thanks{Department of Mathematics, The Hong Kong University of Science and Technology.}%
}
\date{}

\begin{document}

\maketitle

\begin{abstract}
Classical optimization and learning-guided search offer complementary approaches to open
problems in discrete mathematics, but their outputs are difficult to compare when search spaces, validity checks, and stopping rules differ. We present \emph{RadonBench}, a benchmark for constructing maximum-cardinality Radon-free lattice configurations: finite subsets of $\Z^d$ admitting
no integral Radon partition. It remains open whether Radon-free configurations with $11$ points in $\Z^3$ or $21$ points in
$\Z^4$ are possible. If they exist, they would improve the corresponding lower bounds on the integer Radon numbers.
We compare exhaustive extension, greedy local search, PatternBoost, AlphaZero, and
AlphaEvolve. All outputs use one success
predicate: a convex target-sized configuration succeeds exactly when its number of integral
Radon points is zero. PatternBoost and AlphaZero construct verified
Radon-free configurations of sizes $10$ and $20$, one point short of the respective
targets; AlphaZero's best convex target terminals have two integral Radon points in both dimensions.
Exhaustive extension over fixed seeds and declared finite search boxes likewise reaches a
minimum of two integral Radon points. Greedy local search and AlphaEvolve each produce
target-sized near misses with one integral Radon point. No audited run
reaches zero. These results establish reproducible near-miss baselines over explicitly scoped
search domains.

\end{abstract}

\section{Introduction}
\label{sec:intro}

Mathematical discovery is rapidly changing with tools from AI.
Classical enumeration and local
optimization, neural construction, search with learned policies, and evaluator-guided
program evolution have all proved useful for combinatorial discovery
\citep{wagner2021,fawzi2022alphatensor,charton2024patternboost,novikov2025alphaevolve}.
Despite this progress, many basic combinatorial quantities, including numerous Ramsey
numbers, remain beyond the reach of both conventional computation and current AI-assisted
methods~\citep{Radziszowski2011SmallRN,McKay1995R45,Exoo1989ALB}. Here we investigate
another equally difficult problem and use it to benchmark modern search techniques.

Radon's theorem is a fundamental result in convex and discrete geometry: any $d+2$ points
in $\R^d$ admit a bipartition whose convex hulls intersect
\citep{Barany2021CombinatorialConvexity,Barvinok2002CourseConvexity,Gruber2007ConvexDiscreteGeometry}.
Motivated by applications in integer programming and combinatorics, researchers have
studied integer versions of Radon's theorem
\citep{onn1991,Eckhoff1993,deloera2017tverberg,DeLoeraGoaocMeunierMustafa2019}. Here the
points have integer coordinates, and the intersection of the two convex hulls must contain
a lattice point. This gives an
\emph{integral Radon partition}. A finite set in $\Z^d$ with no such partition is
\emph{Radon-free}, and the \emph{integer Radon number} $r(\Z^d)$ is the smallest cardinality for which every lattice point configuration has an integral Radon partition. The exact value in the plane is 
$r(\Z^2)=6$, but it remains unknown in higher dimensions~\citep{onn1991,bezdek2003radon}.
We study whether there is a Radon-free configuration with $11$ points in $\Z^3$ or one with $21$ points in $\Z^4$.

\begin{figure}[t]
  \centering
  \includegraphics[width=\textwidth]{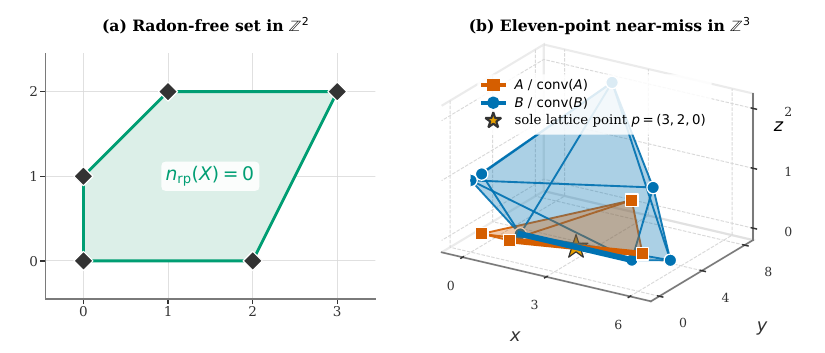}
  \caption{\textbf{The search problem.} (a) Onn's five-point configuration in $\Z^2$ is
  Radon-free~\citep{onn1991}. (b) An $11$-point near-miss in $\Z^3$ found by the declared
  local-search experiment.
  The displayed split has the single shared lattice point $p=(3,2,0)$, hence $\nrp=1$;
  success requires $\nrp=0$.}
  \label{fig:teaser}
\end{figure}

The search for Radon-free configurations offers a compact comparison setting: candidate verification is practical, yet the configuration space is already too large for naive enumeration at the open targets.
Prior work established structural bounds and recognition complexity for integral Radon
partitions~\citep{onn1991,bezdek2003radon}, but those authors did not systematically study constructive search near the open lattice frontiers. 

We introduce \emph{RadonBench}, a benchmark of five method families:
exhaustive extension, greedy local search, a task-adapted PatternBoost solver,
AlphaZero-Radon (our AlphaZero-style solver), and AlphaEvolve-style program evolution.
These methods instantiate two complementary search paradigms. \emph{Cardinality
constructors} grow configurations point by point while maintaining Radon-freeness at
every proper prefix; PatternBoost and AlphaZero-Radon belong primarily to this group.
\emph{Fixed-target optimizers} instead evaluate complete $N$-point configurations and
minimize their number of integral Radon points; this group comprises exhaustive
extension, greedy local search, and AlphaEvolve-style program evolution. AlphaZero-Radon is construction-oriented but
also scores convex $N$-point terminals, so it contributes to the fixed-target comparison
whenever it reaches the horizon.
The methods do not search one common finite domain and their computational units are not
matched. We therefore compare verified outputs rather than claim a ranking of search
efficiency. In particular, we separate two questions:
\begin{enumerate}[leftmargin=1.5em,itemsep=1pt,topsep=2pt]
  \item[\textbf{RQ1}] Does any declared protocol produce a verifier-accepted configuration
  at an open target?
  \item[\textbf{RQ2}] If a run is unsuccessful, what does it nevertheless certify: the
  largest verified Radon-free cardinality $k_{\max}$, or the minimum target-sized
  Radon-point count $r_{\min}(N)$?
\end{enumerate}
This separation prevents a legal $(N-1)$-point prefix from being treated as a scored
$N$-point candidate. Explicit $10$-point and $20$-point witnesses provide finite-box positive
controls immediately below the two frontiers; smaller cardinalities need not be benchmarked
separately because Radon-freeness is hereditary.

Our contributions are twofold. First, we define a common success predicate, graded score,
and reporting contract for search methods with different proposal spaces. Second, we give an
auditable outcome comparison across five classical and learning-guided method families in $\Z^3$
and $\Z^4$, while making each method's reachable configurations and stopping conditions
explicit. Verifier derivations, hyperparameter tables, search-space diagnostics, and
additional experiments are provided in the appendices.

No audited run reported here resolves either open frontier. All negative observations are
limited to the candidates or finite domains searched by the declared protocols; they are not
non-existence results for $\Z^3$ or $\Z^4$.

\section{Problem and Evaluation}
\label{sec:overview}

Let $X\subset\Z^d$ be a finite set of distinct points. An \emph{integral Radon partition}
is a split $X=A\sqcup B$ into nonempty parts satisfying
$\conv(A)\cap\conv(B)\cap\Z^d\ne\varnothing$. We call every lattice point in such an
intersection an integral Radon point and write
\[
  \nrp(X)=\#\{p\in\Z^d:p\text{ is an integral Radon point of }X\}.
\]
Thus $X$ is Radon-free exactly when $\nrp(X)=0$. Every such configuration is in
convex position: if $x\in\conv(X\setminus\{x\})$, then
$\{x\}\sqcup(X\setminus\{x\})$ already has the shared lattice point $x$.
Known global bounds establish the existence of Radon-free sets through sizes $10$ in
$\Z^3$ and $20$ in $\Z^4$, while $N=11$ and $N=21$ are the respective open
targets~\citep{onn1991,bezdek2003radon}. Separate explicit witnesses certify those two
feasible cardinalities immediately below the open targets.

\paragraph{Shared verification.}
Every reported output is judged by the same two-stage evaluator: it first checks convex position and then
enumerates lattice points in $\conv(X)$ and tests whether each supports an integral Radon
partition. A candidate succeeds only if it has the target cardinality, is in convex
position, and has $\nrp=0$. The full evaluator and audit procedure appear in Appendix~\ref{app:verifier}.

For each lattice point $p\in(\conv(X)\cap\Z^d)\setminus X$, the evaluator tests whether
there are weights $\lambda_i\ge0$ and binary assignments $b_i\in\{0,1\}$ satisfying
\[
 \textstyle
 \sum_i b_i\lambda_i x_i=p,\quad
 \sum_i(1-b_i)\lambda_i x_i=p,\quad
 \sum_i b_i\lambda_i=1,\quad
 \sum_i\lambda_i=2.
\]
Here $b_i=1$ assigns $x_i$ to one side of the partition and $b_i=0$ assigns it to the
other. The first two equalities express the same point $p$ as a convex combination of the
two sides. The third equality normalizes the first combination; together with
$\sum_i\lambda_i=2$, it also makes the coefficients of the second combination sum to one.
Thus both sides are nonempty, and feasibility holds exactly when $p$ is an integral Radon
point for some partition. PyNormaliz enumerates the candidate lattice points and Gurobi
performs these split tests~\citep{bruns2010normaliz,gurobi2026}. Each feasible $p$ is
counted once, irrespective of how many partitions witness it, and their number is $\nrp$.
The input points are omitted because the preceding convex-position check guarantees that
none lies in the convex hull of the others.

\begin{table}[t]
\caption{Open targets, known feasible configurations immediately below them, and the
coordinate domain used by each method. Box dimensions are side lengths. ``No finite box''
means that the evaluator accepts arbitrary finite integer coordinates. The method-specific
domains are not box-matched.}
\label{tab:main-instances}
\centering
\small
\setlength{\tabcolsep}{4pt}
\resizebox{\textwidth}{!}{%
\begin{tabular}{@{}lllp{0.34\linewidth}@{}}
\toprule
Instance & Known feasible configuration & Method & Coordinate domain \\
\midrule
\multirow{5}{*}{$\Z^3$, target $11$}
& \multirow{5}{*}{10-point Radon-free}
& Exhaustive extension & Added point: $20\!\times\!20\!\times\!8$ box \\
& & Greedy local search & All points: $20\!\times\!20\!\times\!5$ box \\
& & PatternBoost & All points: $20\!\times\!20\!\times\!5$ box \\
& & AlphaZero-Radon & All points: $20\!\times\!20\!\times\!5$ box \\
& & AlphaEvolve-style & No finite box \\
\addlinespace
\multirow{5}{*}{$\Z^4$, target $21$}
& \multirow{5}{*}{20-point Radon-free}
& Exhaustive extension & Added point: $20\!\times\!20\!\times\!5\!\times\!5$ box \\
& & Greedy local search & All points: $20\!\times\!20\!\times\!5\!\times\!5$ box \\
& & PatternBoost & All points: $6\!\times\!6\!\times\!6\!\times\!6$ box \\
& & AlphaZero-Radon & All points: $6\!\times\!6\!\times\!6\!\times\!6$ box \\
& & AlphaEvolve-style & No finite box \\
\bottomrule
\end{tabular}
}
\end{table}

\paragraph{Output semantics.}
A single leaderboard would conflate variable-length construction with fixed-target repair.
For audited outputs $\mathcal O$ of a declared run, we therefore report
\[
\begin{aligned}
 k_{\max}&=\max\{|X|:X\in\mathcal O,\ X\text{ convex},\ \nrp(X)=0\},\\
 r_{\min}(N)&=\min\{\nrp(X):X\in\mathcal O,\ |X|=N,\ X\text{ convex}\}.
\end{aligned}
\]
The second quantity is undefined if no convex target candidate is emitted. Method-native
counts and layer profiles describe outputs but are not pooled into cross-method success
rates. Search boxes, proposal domains, stopping rules, and budgets are stated per method;
shared verification makes certificates and scores comparable, not search coverage or
efficiency. Appendix~\ref{app:calibration-witnesses} gives the finite-box witnesses, and
Appendix~\ref{app:impl} records the per-method configurations.

\section{Methods}
\label{sec:methods}

All five methods use the target cardinality and verifier from \Cref{sec:overview}. For a
method with an origin-anchored finite coordinate domain of side lengths
$L=(L_1,\ldots,L_d)$, we write
$\Lambda(L)=\prod_{j=1}^{d}\{0,\ldots,L_j-1\}$.
No single box is shared across methods: each protocol has the method-specific coordinate
domain listed in \Cref{tab:main-instances}, and the program-evolution method has no finite coordinate
bound.
Box membership is part of success only when the method itself enforces it.

Exhaustive extension and greedy local search are classical methods; PatternBoost,
AlphaZero-Radon, and AlphaEvolve-style program evolution are learning-guided.
By search objective, PatternBoost and AlphaZero-Radon are primarily cardinality constructors, while exhaustive extension, local search, and
program evolution optimize complete $N$-point configurations. AlphaZero-Radon also scores the terminal
configuration when it reaches $N$. The methods differ in their one-step proposals and
reachable states, summarized in \Cref{tab:search-spaces}; these restrictions scope every
negative result. In \Cref{tab:search-spaces}, ``Starts from'' denotes an externally supplied
configuration; deterministic protocol anchors are specified in the method descriptions.

\begin{table}[H]
\caption{High-level view of the five methods. Gray bands separate the two search
objectives; implementation details appear in the subsections below.}
\label{tab:search-spaces}
\centering
\small
\setlength{\tabcolsep}{3pt}
\renewcommand{\arraystretch}{1.12}
\begin{tabular}{@{}
>{\raggedright\arraybackslash}p{0.15\linewidth}
>{\raggedright\arraybackslash}p{0.25\linewidth}
>{\raggedright\arraybackslash}p{0.50\linewidth}@{}}
\toprule
Method & Starts from & Core idea \\
\midrule
\rowcolor{black!8}
\multicolumn{3}{@{}l@{}}{\strut\textbf{Cardinality construction}: add points while
keeping $\nrp=0$.} \\
PatternBoost
& No points
& Trains a Transformer ordering prior; keeps only convex, Radon-free prefixes.
\\[2pt]
AlphaZero-Radon
& No points
& Uses a set-attention policy and PUCT to add points; scores any convex configuration that
reaches $N$ points under its evaluator. \\
\addlinespace[3pt]
\rowcolor{black!8}
\multicolumn{3}{@{}l@{}}{\strut\textbf{Fixed-target optimization}: minimize $\nrp$
over $N$-point configurations.} \\
Exhaustive extension
& A Radon-free set of $N-1$ points
& Tries every location for one final point in a finite box. \\[2pt]
Greedy local search
& A convex configuration of $N$ points
& Moves one point at a time whenever the number of integral Radon points decreases. \\[2pt]
AlphaEvolve-style
program evolution
& 16 Radon-free examples and a starter program
& Uses an LLM to evolve code that proposes complete $N$-point configurations. \\
\bottomrule
\end{tabular}
\end{table}

\subsection{Shared interface}
\label{sec:interface}

Every method calls the shared convex-position and Radon-point evaluator described in
\Cref{sec:overview}. A proposal may emit a complete configuration or build/edit one
sequentially. Reported outputs use the same success predicate, and every retained coordinate
array is replayed with the reference evaluator.
The exact implementation and the exhaustive driver's equivalent convex-position test are
documented in \Cref{app:verifier,app:exhaustive-config}.

\subsection{Exhaustive add-one-point}
\label{sec:exhaustive}

For each of 50 fixed Radon-free $(N-1)$-point seeds $S$, this baseline tests every
unoccupied position $q$ in a declared finite box $B_d$ and retains convex extensions. Its
per-seed objective is
\begin{equation}
\label{eq:exhaustive-seed-score}
 m_{B_d}(S)=\min_{\substack{q\in B_d\setminus S:\\
 S\cup\{q\}\text{ in convex position}}}
 \nrp\!\left(S\cup\{q\}\right).
\end{equation}
``Exhaustive'' therefore means exhaustive over the final point after fixing $S$ and $B_d$,
not over all seeds or all $N$-point configurations. Exact boxes, coverage counts, and
pseudocode are given in \Cref{tab:main-instances,app:exhaustive-config}.

\subsection{Greedy local search}
\label{sec:local-search}

This deterministic-after-initialization repair baseline starts from a convex $N$-point
extension of a stored Radon-free $(N-1)$-point configuration. A sweep visits points
in stored order and evaluates all full-coordinate radius-one moves
\[
 x_i+\delta,\qquad
 \delta\in\{-1,0,1\}^{d}\setminus\{\mathbf{0}\},
\]
that remain distinct, convex, and inside the method's box. It accepts the admissible move
with the lowest strictly smaller $\nrp$, using a fixed direction order for ties, and stops
at $\nrp=0$, after a non-improving sweep, or after 100 sweeps. Because it changes one point
at a time and rejects plateaus and temporary increases, it cannot make coordinated moves or
cross score barriers. Initialization and ordering details are in \Cref{app:local-config}.

\subsection{PatternBoost generation}
\label{sec:pb}

Our PatternBoost track is an iterative construction--training loop adapted from
\citet{charton2024patternboost}. Configurations in $\Lambda(L)$ are encoded as
mixed-radix point-token sequences. Generation 0 greedily grows empty states by accepting an
addition only when the new prefix is convex and has $\nrp=0$. The largest terminal states
are retained, symmetry-augmented, and used to train a decoder-only
Transformer~\citep{vaswani2017transformer};
sampled sequences seed the next generation and are completed by the same greedy legality
test. The model is therefore trained as an ordering prior, while the verifier enforces every
constraint. Here $\nrp$ is a gate rather than a graded objective: an illegal addition is
rejected, and construction stops when no tested point extends the prefix. Tokenization,
selection, and training details are in \Cref{app:pb-config}.

\subsection{AlphaZero-style construction: AlphaZero-Radon}
\label{sec:az}

AlphaZero-Radon is a task-specific adaptation of the AlphaZero algorithmic
template~\citep{silver2018general}, not a reuse of its chess, shogi, or Go model. It retains
policy--value-guided PUCT and an iterative search--training loop, but replaces the original
game representation and residual convolutional network. Our single-agent construction task
uses unordered point-set states, a permutation-invariant set-attention policy--value
network~\citep{lee2019settransformer}, lattice-cell actions, and verifier-defined outcomes.
Each trajectory starts at the fixed origin and adds the remaining points in $\Lambda(L)$.
The network scores lattice locations, and Monte Carlo tree search plans with the shaped edge
reward
\begin{equation}
\label{eq:reward}
r(s')=
\begin{cases}
0 & \text{if }s'\text{ is not in convex position},\\[3pt]
\dfrac{C_{\mathrm{R}}}{N(1+\nrp(s'))}
  & \text{if }s'\text{ is convex and }\nrp(s')>0,\\[7pt]
\dfrac{C_{\mathrm{S}}}{N}
  & \text{if }|s'|=N,\ s'\text{ is convex, and }\nrp(s')=0,\\[7pt]
\dfrac{C_{\mathrm{step}}}{N}
  & \text{if }|s'|<N,\ s'\text{ is convex, and }\nrp(s')=0 .
\end{cases}
\end{equation}
with $C_{\mathrm{step}}=1$, $C_{\mathrm{R}}=2$, and $C_{\mathrm{S}}=25$. A non-convex or
Radon-positive state terminates the simulated branch; convex Radon-positive states receive
the displayed graded terminal reward at every cardinality. We report fixed-target quality
only for convex terminals at the horizon.

To control branching, each expanded node retains the $K=64$ unoccupied actions with the
largest network probabilities. A child found terminal during search is marked and skipped
by subsequent tree selections. Each decision uses 500 simulations, after which the
trajectory generator samples from the root visit distribution at temperature one. Training,
architecture, and the exact Plan-D search procedure are specified in \Cref{app:az-config}.

\subsection{AlphaEvolve-style program evolution}
\label{sec:alphaevolve}

AlphaEvolve~\citep{novikov2025alphaevolve} treats executable programs, rather than point
sets, as the population under search. We instantiate this with
OpenEvolve~\citep{openevolve2025}: an LLM proposes exact search-and-replace edits to a Python
block, a fixed evaluator executes the edited constructor, and an island-based MAP-Elites
archive~\citep{mouret2015mapelites} retains programs by both fitness and behavioural
features. Thus PatternBoost models a distribution over point sequences, whereas the
AlphaEvolve-style method evolves the algorithm that constructs or repairs point sets.

The immutable wrapper gives each program sixteen fixed Radon-free $(N-1)$-point
references and the verifier. One evaluation permits at most $1{,}000$
\texttt{find\_rps} calls and returns at most 32 integer $N\times d$ candidates. Only four
constructor functions may evolve; the wrapper, resource accounting, geometry code, and
external evaluator remain fixed. The initial program implements the interface but emits no
candidate, so it contributes no hand-written search heuristic.

After the allowance closes, the evaluator validates shapes, finiteness, integrality and
distinctness, then recomputes convexity and $\nrp$ for every returned candidate. Program
fitness is the maximum score of one candidate; statistics from different candidates are
never mixed. A convex target with $k>0$ Radon points scores $0.50+0.49/(1+k)$; a convex
Radon-free target scores $1$. The full fitness, evolution loop, and audit fields are given
in \Cref{app:ae-config}.

\section{Results}
\label{sec:results}

The five methods share one headline question: did the declared run produce a verifier-accepted
configuration at the open target? Table~\ref{tab:main-results} therefore reports target
recovery first. When a run does not recover the target, it keeps two secondary outcomes in
separate columns: $k_{\max}$, the largest fully Radon-free configuration
emitted at any cardinality, and $r_{\min}(N)$, the smallest verifier score among emitted
convex target-sized configurations. These quantities describe different output types and
are not combined into a cross-method ranking.

\begin{table}[H]
\caption{Verified outcomes at the two open targets. \textsc{Found} denotes a verifier-accepted target-sized output; a dash denotes an inapplicable output type.}
\label{tab:main-results}
\centering
\footnotesize
\setlength{\tabcolsep}{5pt}
\renewcommand{\arraystretch}{1.02}
\begin{tabular}{@{}lccc@{}}
\toprule
Method & Found? & $k_{\max}\uparrow$ & $r_{\min}(N)\downarrow$ \\
\midrule
\multicolumn{4}{@{}l}{\textit{$\Z^3$, target $N=11$}} \\
PatternBoost & No & $10$ & -- \\
AlphaZero-Radon & No & $10$ & $2$ \\
Exhaustive extension & No & -- & $2$ \\
Greedy local search & No & -- & $1$ \\
AlphaEvolve-style & No & -- & $1$ \\
\addlinespace[2pt]
\multicolumn{4}{@{}l}{\textit{$\Z^4$, target $N=21$}} \\
PatternBoost & No & $20$ & -- \\
AlphaZero-Radon & No & $20$ & $2$ \\
Exhaustive extension & No & -- & $2$ \\
Greedy local search & No & -- & $1$ \\
AlphaEvolve-style & No & -- & $1$ \\
\bottomrule
\end{tabular}
\end{table}

\begin{figure}[t]
  \centering
  \includegraphics[width=\textwidth]{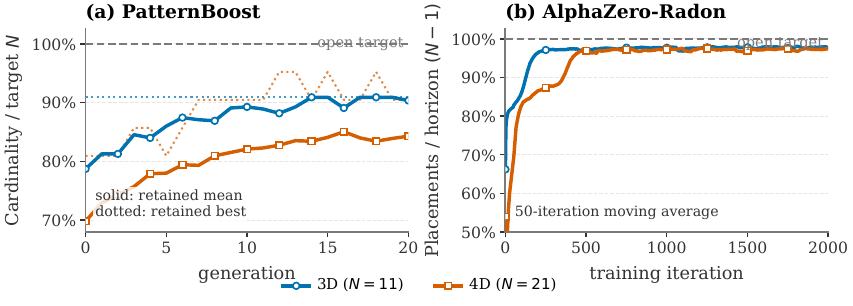}
  \caption{\textbf{Method-native construction progress.}
  (a) For PatternBoost, solid curves show the mean cardinality among the 50 retained
  configurations and dotted curves show the retained maximum at each generation.
  (b) For AlphaZero-Radon, curves show the mean number of post-origin placements per
  self-play trajectory, smoothed with a 50-iteration moving average. Values are normalized
  by the open target cardinality in (a) and by the post-origin horizon $N-1$ in (b).
  The dashed line marks target completion. Horizontal axes are method-specific and do not
  represent matched compute; each curve summarizes one complete declared run.}
  \label{fig:construction-progress}
\end{figure}

\paragraph{RQ1: Did a protocol recover an open-target counterexample?}
No completed PatternBoost run, exhaustive add-one-point experiment, AlphaZero-Radon run,
\SearchLocalStarts-start greedy-local-search run, or AlphaEvolve-style program-evolution run produces a verified
open-target counterexample.

\begin{figure}[t]
  \centering
  \includegraphics[width=\textwidth]{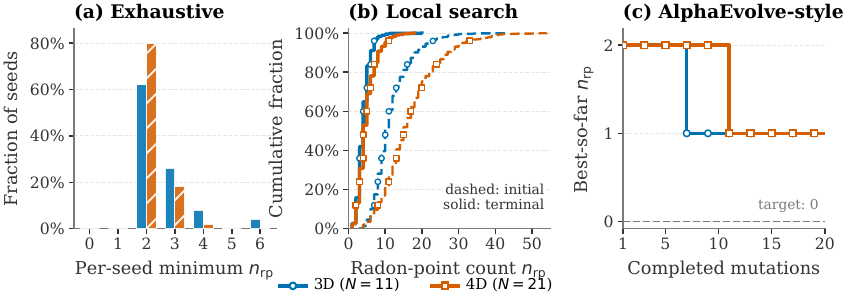}
  \caption{\textbf{Fixed-target search outcomes under the current declared protocols.}
  (a) For exhaustive extension, bars show the distribution of the minimum $\nrp$ obtained
  after searching the full declared box for each of 50 fixed seeds.
  (b) For greedy local search, curves show empirical CDFs of the initial and terminal
  scores over 1,000 selected starts.
  (c) For AlphaEvolve-style program evolution, curves show the best-so-far score over 20
  completed mutations; the first score-one program appears at mutation 7 in 3D and 11 in
  4D. Colors identify the dimension. Seeds, starts, and mutations are protocol-native
  units and do not represent matched compute.}
  \label{fig:fixed-target-outcomes}
\end{figure}

\paragraph{RQ2: What certified output did an unsuccessful run produce?}
PatternBoost reaches $k_{\max}=10$ in $\Z^3$ and $20$ in $\Z^4$. Across its complete
2,000-iteration runs, AlphaZero-Radon reaches $k_{\max}=10$ in 3D and $20$ in 4D; among its
convex target terminals, the minimum score is $2$ in both dimensions. Exhaustive add-one-point search
also attains minimum score $2$ in both dimensions after covering every extension position
in each of its \SearchExhaustiveSeeds\ fixed-seed boxes. Greedy local search attains score $1$ in both
dimensions (on \SearchLocalThreeScoreOne\ of \SearchLocalStarts\ 3D terminals and
\SearchLocalFourScoreOne\ of \SearchLocalStarts\ 4D terminals), but never reaches
zero.
The AlphaEvolve-style method likewise attains $r_{\min}=1$ in both dimensions: its best programs return
convex 11- and 21-point batches whose best individual candidates each have one Radon point.
The two
output types answer different questions: the former certifies a fully valid shorter configuration, whereas the latter
measures a target-sized near miss. We therefore neither order them against one another nor
pool method-native candidate counts into a success rate.
Figure~\ref{fig:construction-progress} shows within-run progress for the two cardinality
constructors; Figure~\ref{fig:fixed-target-outcomes} gives the complementary
fixed-target outcomes for exhaustive extension, greedy local search, and AlphaEvolve-style
program evolution. Both use method-native axes and do not compare computational efficiency.
Appendix~\ref{app:ai-3d-configs} provides representative verified constructions.

The experiments are not compute- or verifier-query-matched. Method-native run lengths and
budgets are reported for reproducibility, but Table~\ref{tab:main-results} compares
verified outputs rather than search efficiency. Appendix~\ref{sec:experiments} records the
completed runs and retained outputs.

\section{Related Work}
\label{sec:related}

Onn formalized integral Radon partitions in the lattice setting, proved
$r(\Z^2)=6$, established general bounds, and studied recognition complexity~\citep{onn1991}.
Bezdek and Blokhuis proved the upper bound $r(\Z^3)\le17$~\citep{bezdek2003radon}; integral
Radon partitions are also integer $2$-Tverberg partitions
\citep{deloera2017tverberg}. These works motivate the open frontiers but do not provide a
multi-method empirical study of constructive search.

Relevant ingredients in machine-assisted mathematical discovery include neural
construction~\citep{wagner2021}, policy--value-guided tree search from
AlphaZero~\citep{silver2018general}, its task-specific adaptation to algorithm discovery in
AlphaTensor~\citep{fawzi2022alphatensor}, Transformer-guided search
\citep{charton2024patternboost}, and LLM program evolution
\citep{romera2024funsearch,novikov2025alphaevolve}. A recent comparative precedent studies
classical, PatternBoost, and reinforcement-learning methods for the
no-three-in-line problem~\citep{ramanathan2025three}. RadonBench instead uses a
lattice-convexity verifier and explicitly separates constructive from fixed-cardinality
outputs. Complete computational resolutions of combinatorial existence problems require
exhaustive search and checkable certificates~\citep{heule2016pythagorean,heule2018schur,
brakensiek2020keller}; our bounded searches do not provide such a non-existence certificate.

\section{Discussion}
\label{sec:discussion}

Neither PatternBoost, exhaustive extension, nor AlphaZero-Radon reaches an open target; greedy
local search likewise finds no target across \SearchLocalStarts\ starts per dimension but
reaches score-one near misses in both. Two explanations remain inseparable: the declared searches may cover too little of
their method-specific domains, or a Radon-free configuration may not exist at that
cardinality. These observations support only search-space diagnoses. A constructor's
legal-prefix frontier reflects its proposal distribution and cutoff, not a global
obstruction. After \SearchOEMutations\ mutations, the AlphaEvolve-style runs in both dimensions reach score-one
near misses.

\paragraph{Reading the results through each search space.}
Exhaustive extension provides a finite-domain completeness statement only after fixing a
legal $(N-1)$-point seed and a finite box for the last point. The completed experiment
visits every position in each declared box, but its negative result concerns only those \SearchExhaustiveSeeds\
fixed seeds per dimension, not configurations that require changing a seed or adding a
point outside the box.

Greedy local search has the opposite search-space design. It can modify every coordinate of
every point of an $N$-point configuration over successive sweeps, but each proposal moves
only one point by Chebyshev radius one, remains inside the fixed box, and must immediately
lower the integer-valued score. All 2,000 reported runs stall at positive score, including
\SearchLocalThreeScoreOne\ of \SearchLocalStarts\ in 3D and
\SearchLocalFourScoreOne\ of \SearchLocalStarts\ in 4D at score one. A missing path may therefore require coordinated changes, a neutral or uphill
intermediate state, or leaving the declared box.

PatternBoost and AlphaZero-Radon both build point sets sequentially. PatternBoost assigns no
graded score to a failing prefix, whereas AlphaZero-Radon gives a larger terminal reward to a
convex Radon-positive state with fewer integral Radon points. PatternBoost fits an ordering distribution
from previously retained configurations and then uses the verifier to accept or skip sampled
points. AlphaZero-Radon uses its policy to retain a top-$K$ action set at each expanded node and
spends MCTS simulations only within that restricted support. A frontier at $N-1$ therefore
means that the learned proposal mechanism did not produce a verified continuation under the
declared run; it is not evidence that no continuation exists elsewhere in the box.

AlphaEvolve-style program evolution has the broadest coordinate interface because an evolved program may emit
arbitrary finite integer coordinates. Its practical search space is nevertheless the set of
programs reached by its declared mutation lineage and the bounded batch of candidates each
program returns. Its 3D and 4D lineages both attain one-Radon-point target configurations,
but neither returns a Radon-free target. These distinctions are why we keep
target recovery, valid-prefix cardinality, and target-sized score in separate table columns
rather than treating the five methods as an efficiency leaderboard.

\paragraph{What defines a reproducible search-space claim.}
A finite-box exhaustive search must specify its fixed seeds and coordinate box; a repair
method must specify its initialization, move radius, number of points changed together, and
acceptance rule. A learned constructor additionally needs its action vocabulary, candidate
truncation, and terminal handling, while program evolution needs its evaluator batch and
number of mutations. Omitting any of these choices changes the reachable candidates. We expose
them so future runs can vary one restriction at a time without changing the success
predicate. The same accounting explains why a grid position, a repair trajectory, an MCTS
game, and a program-returned candidate are not interchangeable trials.

A complete negative result would require more than exhaustive evaluation inside one
coordinate box. One would need a proved finite representative family, or an invariant
coordinate bound, for lower-cardinality Radon-free configurations and a proved finite
extension region for every representative. Affine unimodular transformations preserve Radon-freeness but need not preserve a box, so bounded-box exhaustion alone does not cover all
equivalence classes. The appendices state the exact scope of every filled result.

\paragraph{Implications for search design.}
The framework suggests three protocol-level choices for future runs. First, repair methods
should test coordinated moves or permit neutral and uphill transitions instead of enlarging
the coordinate radius alone. Second, initialization and box constraints deserve the same
care as the stopping rule because both restrict reachable basins. Third,
evaluation should retain the objective each algorithm actually optimizes. Constructors
should report the largest verified legal prefix, whereas fixed-target methods should report
the best verified target score; neither replaces the common binary target-recovery outcome.

\paragraph{Limitations.}
The methods use different proposal domains, move sets, and run lengths; hardware
and verifier calls are not matched. PatternBoost has one 20-generation run per dimension,
and AlphaZero-Radon has one complete 2,000-iteration cold-start run per dimension.
AlphaEvolve-style program evolution has one \SearchOEMutations-mutation lineage per dimension rather than repeated independent trials.
The greedy-local-search results are limited to \SearchLocalStarts\ fixed starts per dimension, strict single-point
descent, and dimension-specific fixed boxes. The exhaustive experiment fixes only \SearchExhaustiveSeeds\ catalog seeds per dimension
and cannot exclude extensions of other seeds or points outside its declared boxes.
Consequently, Table~\ref{tab:main-results} compares verified outputs, not sample or
computational efficiency. Although the two constructive methods share the
$6\times6\times6\times6$ box in 4D,
their search distributions and budgets are not matched.

\section{Conclusion}
\label{sec:conclusion}

RadonBench organizes five classical and learning-guided searches around a common target-recovery
criterion while preserving their different output semantics. PatternBoost reaches the known
calibration cardinalities $10$ and $20$; across its complete 2,000-iteration runs,
AlphaZero-Radon reaches the same cardinalities and target score $2$ in both dimensions, while
exhaustive add-one-point search obtains
target score $2$ but no target recovery after completely visiting the declared box for each
of \SearchExhaustiveSeeds\ fixed seeds in both dimensions. Full-coordinate greedy local search reaches score
$1$ but no target across \SearchLocalStarts\ starts per dimension. AlphaEvolve-style program evolution also reaches score $1$ in
both dimensions after \SearchOEMutations\ mutations, without
recovering either target. These results do not resolve
the integer Radon numbers, but the success-first reporting scheme provides a reproducible
framework for adding future searches without conflating solution quality with compute
efficiency.

\section*{Acknowledgments}
This work was supported in part by NVIDIA through the NVIDIA Academic Grant Program
under the project ``AI Foundation Models for Reasoning in Polyhedral Geometry.''

\clearpage
\phantomsection\label{page:references}
\bibliography{references}
\bibliographystyle{plainnat}

\appendix
\raggedbottom
\setlength{\abovedisplayskip}{7pt plus 2pt minus 2pt}
\setlength{\belowdisplayskip}{7pt plus 2pt minus 2pt}
\setlength{\abovedisplayshortskip}{4pt plus 2pt minus 1pt}
\setlength{\belowdisplayshortskip}{5pt plus 2pt minus 1pt}
\setlength{\jot}{2pt}
\section{Detailed Experimental Outcomes}
\label{sec:experiments}

All declared runs use the stopping rules and sample counts specified in the method appendix.
The headline outcomes appear in Table~\ref{tab:main-results}; this section records only
method-native evidence that is not shown there.

\subsection{Constructive methods}

\paragraph{PatternBoost.}
Both 20-generation loops completed. In the \(20\times20\times5\) box, the largest verified
Radon-free cardinality is \(10\); generation 20 retains 47 ten-point and three nine-point
states. In the \(6\times6\times6\times6\) box, the largest cardinality is \(20\), reached
in generations 12, 13, 15, and 18; generation 20 retains eight nineteen-point,
19 eighteen-point, and 23 seventeen-point states. Neither run reaches its open target.

\paragraph{AlphaZero-Radon.}
Each 2,000-iteration run generates 200,000 trajectories. In 3D, 157,485 trajectories reach
the 11-point horizon and 8,651 of those terminals are convex; their minimum score is
\(r_{\min}(11)=2\). In 4D, 120,973 trajectories reach the 21-point horizon and 2,302 are
convex; their minimum score is \(r_{\min}(21)=2\). The largest verified Radon-free prefixes
have 10 and 20 points, respectively, and neither run reaches its open target.

\subsection{Fixed-target methods}

\paragraph{Exhaustive extension.}
For each dimension, the experiment fixes \SearchExhaustiveSeeds\ Radon-free seeds and tests
every unoccupied lattice position in the declared box. The 3D run tests
\SearchExhaustiveThreeTested\ additions and scores \SearchExhaustiveThreeConvex\ convex
targets; the 4D run tests \SearchExhaustiveFourTested\ additions and scores
\SearchExhaustiveFourConvex\ convex targets. The minimum score is two in both dimensions.
This is exhaustive only for the fixed seed--box pairs.

\begin{table}[t]
\caption{Coverage of the declared exhaustive add-one-point domains. Tested excludes
positions occupied by the fixed seed; convex targets are the extensions scored by the
verifier.}
\label{tab:exhaustive-results}
\centering
\small
\setlength{\tabcolsep}{5pt}
\begin{tabular}{lrrr}
\toprule
Instance & Seeds & Tested & Convex targets \\
\midrule
\(\Z^3,\ N=11\) & \SearchExhaustiveSeeds & \SearchExhaustiveThreeTested & \SearchExhaustiveThreeConvex \\
\(\Z^4,\ N=21\) & \SearchExhaustiveSeeds & \SearchExhaustiveFourTested & \SearchExhaustiveFourConvex \\
\bottomrule
\end{tabular}
\end{table}

\paragraph{Greedy local search.}
Both dimensions use fixed \SearchLocalStarts-start evaluation sets. In 3D,
\SearchLocalThreeImproved\ starts improve and \SearchLocalThreeScoreOne\ terminate at score
one; the median score falls from 10 to 4. In 4D, \SearchLocalFourImproved\ starts improve
and \SearchLocalFourScoreOne\ terminate at score one; the median falls from 15 to 4.
Neither run reaches zero. The reported 4D aggregate is fixed to indices 0--999.

\paragraph{AlphaEvolve-style program evolution.}
Each dimension has one \SearchOEMutations-mutation lineage. Every completed mutation returns
\SearchOECandidatesPerMutation\ convex target-sized configurations under a shared allowance
of \SearchOECalls\ internal \texttt{find\_rps} calls, for \SearchOECandidates\ externally
scored candidates per dimension. Neither lineage reaches zero. The first score-one program
appears at mutation \SearchOEThreeFirstBest\ in 3D and mutation
\SearchOEFourFirstBest\ in 4D.

\section{Additional Experiment Configurations}
\label{app:newconfigs}

\paragraph{Software.} Python~3.11.15, NumPy~2.4.6, PyNormaliz~2.23, Gurobi~13.0.2. The
structural analyses use no GPU; all analysis runs on CPU on a shared machine, so wall-clock
figures are
indicative rather than compute-matched.

\subsection{Finite-box calibration witnesses}
\label{app:calibration-witnesses}

The following ordered coordinate arrays are finite-box positive controls. The 3D witness
has been translated so that its coordinate-wise minima are zero; the 4D witness is a
verified 20-point prefix emitted by the reported AlphaZero-Radon run at iteration 190. The shared
verifier confirms row distinctness, convex position, and $\nrp=0$ for both arrays.

\paragraph{Three-dimensional witness.}
\[
 W_3=\begin{bmatrix}
3&5&0\\8&4&0\\6&2&0\\5&2&0\\7&5&0\\
6&6&1\\6&8&1\\2&1&1\\3&0&1\\0&5&2
\end{bmatrix}
 \subset\Lambda(20,20,5).
\]
This array is the $\Z^3$ positive-control witness used at $n=10$.

\paragraph{Four-dimensional witness.}
\[
 W_4=\begin{bmatrix}
0&0&0&0\\1&0&0&0\\1&0&1&5\\2&0&1&0\\2&0&1&5\\
1&0&0&5\\0&5&0&0\\2&5&1&3\\1&3&1&0\\1&5&0&5\\
0&5&0&5\\1&2&1&5\\1&5&1&4\\2&5&1&1\\2&2&1&5\\
1&5&1&3\\0&1&0&5\\1&4&0&0\\0&0&0&4\\1&5&0&1
\end{bmatrix}
 \subset\Lambda(6,6,6,6).
\]
This array is the $\Z^4$ positive-control witness used at $n=20$.

Every subset of a convex-position configuration remains in convex position, and Radon-freeness is downward closed under taking subsets. Consequently,
auditing the two displayed arrays also certifies feasibility at every smaller cardinality in
the same boxes. The main experiments nevertheless use only $n=10$ and $n=20$ as positive
controls, because smaller subsets add no independent feasibility question. The witnesses
certify feasibility only; they do not imply that a particular solver can reach them from its
initialization.

\subsection{Three-dimensional diversity of learning-guided 10-point configurations}
\label{app:ai-3d-configs}

\begin{figure}[t]
  \centering
  \includegraphics[width=\textwidth]{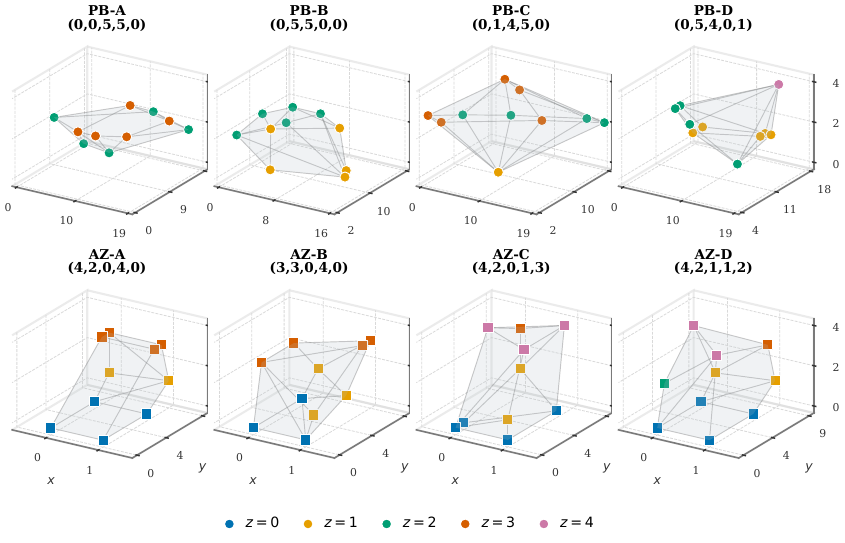}
  \caption{Representative learning-guided 10-point configurations in 3D.
  PB-A--PB-D (circles) are PatternBoost outputs; AZ-A--AZ-D (squares) are
  AlphaZero-Radon construction prefixes. Color encodes height $z$, and each heading gives
  the layer-count vector from $z=0$ to $z=4$. Gray edges and translucent faces
  show the convex-hull scaffold. To keep both compact AlphaZero-Radon configurations
  and box-spanning PatternBoost configurations legible, panels use their own
  displayed $x$--$y$ limits, with the true limits shown by the ticks; the $z$ axis
  is visually expanded. Every panel is a distinct Radon-free configuration. The
  examples illustrate structural diversity
  rather than empirical profile frequencies.}
  \label{fig:ai-3d-configs}
\end{figure}

\Cref{fig:ai-3d-configs} gives eight verified 10-point configurations in
$\Lambda(20,20,5)$. PB-A--PB-D are retained outputs from PatternBoost generation 20;
AZ-A--AZ-D are retained AlphaZero-Radon construction prefixes from iterations 1961, 1962,
and 1999. The panel headings report the layer profiles as counts on
$z=0,\ldots,4$. All eight profiles are distinct:
$(0,0,5,5,0)$, $(0,5,5,0,0)$, $(0,1,4,5,0)$, $(0,5,4,0,1)$,
$(4,2,0,4,0)$, $(3,3,0,4,0)$, $(4,2,0,1,3)$, and $(4,2,1,1,2)$.
The examples were deliberately selected to expose different layerwise structures,
not to estimate how frequently either method produces a profile. The reference verifier
independently confirms convex position and $\nrp=0$ for every displayed configuration.

\subsection{Result aggregation and audit}
\label{app:best-candidate-selection}

For constructive protocols, we replay retained serialized terminal states and take the
largest cardinality passing convexity and $\nrp=0$; this produces $k_{\max}$ in
\Cref{tab:main-results}. For fixed-target protocols, we discard repeated-row or
non-convex arrays and take the minimum $\nrp$ among the remaining $N$-point candidates;
this produces $r_{\min}(N)$ in \Cref{tab:main-results}. Ties may be resolved by
lexicographic order of the sorted integer rows when an exact representative is needed.
Each reported result records its source run, iteration or generation, verifier output, and
artifact path. AlphaZero-Radon aggregation includes all 2,000 completed iterations in each
dimension.

\section{Implementation details}
\label{app:impl}

This appendix provides pseudocode for the reference verifier and every evaluated search
method. The complete algorithmic specifications appear in
\Cref{alg:reference-verifier,alg:exhaustive-add-one,alg:local-search,alg:patternboost,alg:alphaevolve-radon,alg:alphazero,alg:az-puct}.

\subsection{Reference verifier}
\label{app:verifier}

The benchmark verifier consists of the functions \texttt{is\_conv\_position\_all()} and
\texttt{find\_rps()}. For $X=\{x_1,\ldots,x_n\}\subset\Z^d$, the first routine performs
one leave-one-out linear-program feasibility test per input point. For each $i$, it asks
whether coefficients $\alpha_j\ge0$ exist such that
\begin{equation}
  \sum_{j\ne i}\alpha_j=1,
  \qquad
  \sum_{j\ne i}\alpha_j x_j=x_i.
  \label{eq:convex-position-lp}
\end{equation}
Feasibility means $x_i\in\conv(X\setminus\{x_i\})$ and the candidate is rejected;
\texttt{is\_conv\_position\_all()} returns \texttt{True} exactly when every leave-one-out
model is infeasible.

The function \texttt{find\_rps()} first passes the homogenized polytope to PyNormaliz
~\citep{bruns2010normaliz} and reads \texttt{Deg1Elements}, which enumerates the lattice
points in $\conv(X)$. Input vertices are removed. For every remaining lattice point
$p$, the verifier asks whether there are $\lambda_i\ge0$ and binary $b_i\in\{0,1\}$ such
that
\begin{align}
 \sum_i b_i\lambda_i x_i &= p, &
 \sum_i (1-b_i)\lambda_i x_i &= p,\nonumber\\
 \sum_i b_i\lambda_i &=1, &
 \sum_i\lambda_i &=2.
 \label{eq:radon-verifier}
\end{align}
The variable $b_i$ assigns $x_i$ to the first part when $b_i=1$ and to the second part
when $b_i=0$; the corresponding products select the weights used by each convex
combination. The first coefficient sum is constrained to one, and the second also sums to
one because
$\sum_i(1-b_i)\lambda_i=\sum_i\lambda_i-\sum_i b_i\lambda_i=2-1$.
Thus a feasible solution places $p$ in the convex hulls of two nonempty disjoint parts.
Conversely, any integral Radon partition and an integral Radon point give a feasible assignment. The
implementation solves this model with
Gurobi~\citep{gurobi2026}, sets \texttt{NonConvex=2} for the binary--continuous products,
and tests each candidate $p$. It returns each feasible lattice point once, irrespective of
the number of partitions that witness it; hence the
benchmark score is $\nrp(X)=\lvert\texttt{find\_rps}(X)\rvert$. A candidate is a verified
Radon-free configuration exactly when \texttt{is\_conv\_position\_all()} returns \texttt{True} and
\texttt{find\_rps()} is empty.

\begin{algorithm}[H]
\caption{Reference convexity and integral-Radon-point evaluator.}
\label{alg:reference-verifier}
\begin{algorithmic}[1]
\Require finite set of distinct lattice points $X=\{x_1,\ldots,x_n\}\subset\Z^d$
\Ensure convex-position flag $c$ and set $R$ of integral Radon points
\For{$i=1,\ldots,n$}
  \If{$x_i\in\conv(X\setminus\{x_i\})$ according to \Cref{eq:convex-position-lp}}
    \State \Return $(\textsc{False},\varnothing)$
  \EndIf
\EndFor
\State $P\gets(\conv(X)\cap\Z^d)\setminus X$ using PyNormaliz
\State $R\gets\varnothing$
\For{$p\in P$}
  \If{the feasibility system in \Cref{eq:radon-verifier} accepts $p$}
    \State $R\gets R\cup\{p\}$
  \EndIf
\EndFor
\State \Return $(\textsc{True},R)$
\end{algorithmic}
\end{algorithm}

Following the canonical implementation, \textsc{Infeasible} rejects a candidate Radon
point and every other returned status counts it. All complete Radon-count evaluations use
this same rule.

\paragraph{Cross-track accounting.} The five tracks are not compute-matched.
Method-native stopping rules are part of each protocol, but raw verifier-call totals are
not used as a common budget or efficiency metric. Reported local-search aggregates are
computed from retained JSON/NPZ artifacts for both the 3D and 4D instances; retained outputs
likewise support both dimensional instances of the other tracks. A stochastic run is
successful only if a serialized candidate passes the common verifier.

The available implementations use the software versions listed
below and in \Cref{app:local-config}. Their evaluator implementations share the formulation
and status rule above. Retained configurations and aggregate logs support the reported
audits.
\paragraph{PatternBoost and AlphaZero-Radon environment.} PatternBoost and AlphaZero-Radon use Python~3.11.15,
PyTorch~2.12.0 with CUDA~13.0, NumPy~2.4.6, SciPy~1.17.1 (including Qhull),
PyNormaliz~2.23, and Gurobi~13.0.2. Neural-network training and generation use one
NVIDIA A100-SXM4 GPU with 80\,GB memory; CPU processes execute the geometric oracle.
We do not request PyTorch's bitwise-deterministic GPU kernels.

\paragraph{Local and exhaustive environment.} These CPU-only searches use Python~3.12.11,
NumPy~1.26.4, SciPy~1.16.2, PyNormaliz~2.19, and Gurobi~12.0.3; the local-search environment
also contains PyTorch~2.8.0 and trimesh~4.8.3. They run on an Intel Xeon Gold 5320 node
(two 26-core sockets, SMT2, 503\,GiB RAM). Local tasks request 32 CPUs and 64\,GB with at
most three concurrent tasks; exhaustive tasks request 8 CPUs and 24\,GB and run one at a
time. Neither requests a GPU.

\paragraph{Program-evolution environment.} OpenEvolve uses Python~3.12.11, OpenEvolve~0.2.27,
NumPy~1.26.4, SciPy~1.16.2, PyNormaliz~2.19, and Gurobi~12.0.3 on the same CPU node. Each
job requests four CPUs, 64\,GB RAM, and no GPU. The method-specific subsections give the
remaining seeds, stopping rules, provenance, and run settings.

\subsection{Exhaustive add-one-point experiment configuration}
\label{app:exhaustive-config}

\paragraph{Dimension-general procedure.}
For a fixed seed $S\subset\Z^d$ and integer box
$B=\prod_{j=1}^d\{\ell_j,\ldots,u_j\}$, the driver visits every $q\in B$ in
deterministic Cartesian-product order. It skips $q\in S$, rejects extensions outside convex
position, and evaluates $\nrp(S\cup\{q\})$ for every remaining extension. It does not stop
when a zero or a new minimum is found. Therefore every unoccupied position in $B$ is either
rejected by the convex-position predicate or assigned a Radon-point count. This is the sense
in which the search is exhaustive. The seed is never modified, and no claim is made about
other seeds or positions outside $B$.

\begin{algorithm}[t]
\caption{Exhaustive add-one-point search within a fixed box}
\label{alg:exhaustive-add-one}
\begin{algorithmic}[1]
\Require verified Radon-free seed $S\subset\Z^d$ with $|S|=N-1$;
         bounds $(\ell_j,u_j)_{j=1}^d$
\Ensure minimum score $m$, minimizers $Q$, score histogram $H$, coverage counts
\State $B\gets\prod_{j=1}^d\{\ell_j,\ldots,u_j\}$; $m\gets+\infty$;
       $Q\gets\varnothing$; $H\gets\varnothing$; $c_{\rm convex}\gets0$
\For{$q\in B$ in Cartesian-product order}
  \If{$q\in S$}
    \State \textbf{continue}
  \EndIf
  \State $X\gets S\cup\{q\}$
  \If{$X$ is not in convex position}
    \State \textbf{continue}
  \EndIf
  \State $c_{\rm convex}\gets c_{\rm convex}+1$
  \State $r\gets\nrp(X)$; $H[r]\gets H[r]+1$
  \If{$r<m$}
    \State $m\gets r$; $Q\gets\{q\}$
  \ElsIf{$r=m$}
    \State $Q\gets Q\cup\{q\}$
  \EndIf
\EndFor
\State \Return $(m,Q,H,|B|-|S|,c_{\rm convex})$
\end{algorithmic}
\end{algorithm}

\paragraph{Frozen seeds and sampling.}
Before evaluation, the experiment fixes \SearchExhaustiveSeeds verified Radon-free seeds per
dimension using NumPy seed $20260819$. The 3D seeds lie in
$\{-10,\ldots,9\}^2\times\{0,\ldots,7\}$; the 4D seeds lie in
$\{0,\ldots,19\}^2\times\{0,\ldots,4\}^2$. Seed selection is independent of
extension scores, and no sampling occurs during a task.

\paragraph{Boxes and coverage.}
The 3D box is
$B_3=\{-10,\ldots,9\}\times\{-10,\ldots,9\}\times\{0,\ldots,7\}$,
with $3{,}200$ positions. Each 10-point seed is contained in $B_3$, leaving $3{,}190$
attempted additions per seed and $\SearchExhaustiveThreeTested$ overall. The 4D box is
$B_4=\{0,\ldots,19\}\times\{0,\ldots,19\}\times\{0,\ldots,4\}^2$,
with $10{,}000$ positions. Each 20-point seed is contained in $B_4$, leaving $9{,}980$
attempted additions per seed and $\SearchExhaustiveFourTested$ overall.

\paragraph{Convex-position and Radon tests.}
Before enumeration, the driver checks that the seed is row-distinct, in convex position,
and has $\nrp(S)=0$. For each extension $X$, it constructs the homogenized polytope with
PyNormaliz and reads \texttt{ExtremeRays}. The extension passes exactly when the returned
rays equal $\{(x,1):x\in X\}$, so every input point is a vertex. This exact rational
extreme-ray test is mathematically equivalent to the reference leave-one-out convex-position
predicate, although it is a different implementation. For every passing extension,
\texttt{find\_rps\_nd()} uses PyNormaliz \texttt{Deg1Elements} and the same Gurobi
partition-feasibility test in \Cref{app:verifier}. Gurobi uses one thread per model; up to
eight lattice points are tested concurrently within a task.

The supplementary manifest records the fixed seed coordinates, declared boxes, and coverage
counts needed to reproduce the reported aggregate.

\subsection{Greedy local-search experiment configuration}
\label{app:local-config}

\paragraph{Dimension-general procedure.}
For an instance $(d,N,L)$, the descent moves all coordinates. Its ordered neighbourhood is
\[
 \mathcal D_d=\{-1,0,1\}^{d}\setminus\{\mathbf{0}\}.
\]
For a box $\Lambda(L)$, proposals outside the box are rejected. The evaluated 3D and 4D
neighbourhoods contain 26 and 80 offsets, respectively, before boundary and geometry
filtering. The method uses
a strict-improvement rule, stored point order, lexicographic direction order, a 100-sweep
cap, and early stopping at score zero or after a full non-improving sweep.

\paragraph{Initial configurations.}
In 3D, initialization adds one uniformly sampled point in
$\{-10,\ldots,9\}^2\times\{0,\ldots,4\}$ to a verified 10-point seed and
accepts the first distinct, convex extension within 100 draws. From the eligible pool,
\SearchLocalStarts starts are sampled with NumPy seed $20260819$ and translated into the
$20\times20\times5$ box. In 4D, one uniformly sampled in-box point is added to each
verified 20-point seed, again requiring distinctness and convex position; the reported set is
the fixed first \SearchLocalStarts starts generated with NumPy seed $20260815$. Initialization
does not filter or rank candidates by $\nrp$.
\Cref{alg:local-search} gives the subsequent deterministic descent. The point order is the
stored array order. Directions are the lexicographically ordered Cartesian product
$\{-1,0,1\}^d$ with the zero vector removed. Only strict improvement is admissible, so the
first direction attaining a given best score wins a tie and equal-score plateaus are never
traversed. No anchoring translation occurs during descent. All accepted 3D configurations
remain inside $\{0,\ldots,19\}^2\times\{0,\ldots,4\}$, and all accepted 4D configurations
remain inside $\{0,\ldots,19\}^2\times\{0,\ldots,4\}^2$.

\begin{algorithm}[t]
\caption{Dimension-general ordered greedy local search.}
\label{alg:local-search}
\begin{algorithmic}[1]
\Require ordered convex configuration $X=(x_1,\ldots,x_N)\subset\Z^d$, sweep cap $T=100$,
ordered directions $\mathcal{D}_d$, score $\nrp$
\For{$t=1,\ldots,T$}
  \State $s_{\mathrm{start}}\gets\nrp(X)$
  \For{$i=1,\ldots,N$}
    \State $s^\star\gets\nrp(X)$; $X^\star\gets X$
    \For{$\delta\in\mathcal{D}_d$ in lexicographic order}
      \State $Y\gets X$ with $x_i\gets x_i+\delta$
      \If{$Y\subseteq\Lambda(L)$, has distinct points, and is in convex position}
        \State $s\gets\nrp(Y)$
        \If{$s<s^\star$} \State $(X^\star,s^\star)\gets(Y,s)$ \EndIf
        \If{$s^\star=0$} \State \textbf{break} \EndIf
      \EndIf
    \EndFor
    \State $X\gets X^\star$
  \EndFor
  \If{$\nrp(X)=0$ or $\nrp(X)\ge s_{\mathrm{start}}$} \State \textbf{break} \EndIf
\EndFor
\State \Return $X$
\end{algorithmic}
\end{algorithm}

\paragraph{Convexity and score implementation.}
For every candidate, convex position is checked point by point.  To test point $x_i$, a
Gurobi feasibility LP asks whether nonnegative coefficients on $X\setminus\{x_i\}$ sum
to one and reproduce $x_i$; feasibility rejects the candidate.  For the objective,
PyNormaliz enumerates the degree-one lattice points of the homogenized polytope and the
implementation removes the input vertices.  For each remaining lattice point $p$, Gurobi
solves the feasibility model
\begin{align*}
 \sum_i b_i\lambda_i x_i &= p, &
 \sum_i (1-b_i)\lambda_i x_i &= p,\\
 \sum_i b_i\lambda_i &=1, &
 \sum_i\lambda_i &=2,
\end{align*}
with $\lambda_i\ge0$ continuous and $b_i\in\{0,1\}$; the model uses
\texttt{NonConvex=2}.  The number of feasible $p$ is the descent score; any solver status
other than \textsc{Infeasible} is treated as feasible.  No Gurobi feasibility,
integrality, or optimality parameters are overridden.  Radon points are memoized within a configuration after its first score evaluation. The fixed selection manifests identify the 1,000 reported starts in each dimension, and the retained trajectories reproduce the aggregate outcomes in Appendix~\ref{sec:experiments}. The CPU implementation uses Python~3.12.11, PyNormaliz~2.19, Gurobi~12.0.3, and SciPy~1.16.2.

\begin{table}[H]
\caption{Greedy local-search hyperparameters shared across dimensions.}
\label{tab:local-hparams}
\centering
\footnotesize
\setlength{\tabcolsep}{3pt}
\begin{tabular}{@{}>{\raggedright\arraybackslash}p{0.53\linewidth}
>{\raggedright\arraybackslash}p{0.43\linewidth}@{}}
\toprule
Parameter & Value \\
\midrule
Initialization draw cap & $100$ per extension request \\
Evaluated starts / seeds & $\SearchLocalStarts$ per dimension / 3D $20260819$, 4D $20260815$ \\
Neighbourhood & $3^d-1$ full-coordinate radius-$1$ moves (3D: 26; 4D: 80) \\
Point / direction order & stored index / lexicographic \\
Acceptance / plateaus & strict $\nrp$ decrease / disallowed \\
Coordinate bounds during descent & 3D: $[0,19]^2\times[0,4]$; 4D: $[0,19]^2\times[0,4]^2$ \\
Maximum sweeps / early stop & $100$ / full non-improving sweep or $\nrp=0$ \\
\bottomrule
\end{tabular}
\end{table}

\subsection{PatternBoost experiment configuration}
\label{app:pb-config}

For an instance $(d,N,L)$, PatternBoost uses variable-length add-point sequences in
$\Lambda(L)$. Its vocabulary has $1+\prod_{j=1}^{d}L_j$ entries: one shared
\textsc{Start}/\textsc{Stop} token and one token per lattice point. A sequence has the form
$[0,\tau_L(p_1),\ldots,\tau_L(p_k),0]$, where the points are sorted lexicographically
before supervised training; losses after the first stop token are masked. The reference
incremental legality test accepts a prefix only when it is in convex position and has no
integral Radon point. This is stricter than merely filtering complete model outputs: every
training configuration and every saturated output is legal at each accepted prefix.
All dimension-independent architectural, optimization, selection, generation, and
saturation hyperparameters are shared across dimensions.

The completed three-dimensional run instantiates $L=(20,20,5)$ and $N=11$, giving a
$2{,}001$-token vocabulary. The completed four-dimensional run uses $L=(6,6,6,6)$ and
$N=21$, giving a $1{,}297$-token vocabulary.

\begin{algorithm}[H]
\caption{PatternBoost with verifier-gated incremental saturation.}
\label{alg:patternboost}
\begin{algorithmic}[1]
\Require box $L$, maximum size $N$, generations $G$, state-batch size $B$, keep fraction
$q$, target retained-set size $T$, reference predicate \Call{Legal}{S}
\State form $T/q$ independent empty seeds; saturate each seed with \Call{Legal}{}
\State retain the top $qB$ terminal states within each batch to obtain $\mathcal{D}_0$
\For{$i=0,\ldots,G-1$}
  \State split $\mathcal{D}_i$ by complete symmetry orbit into train/validation sets
  \State augment both splits by box-preserving reflections and axis permutations
  \State train a next-token Transformer; retain the lowest-validation-loss checkpoint
  \State sample $T/q$ unconstrained token sequences from \textsc{Start}
  \State decode each sequence in order, skipping a token unless it denotes an unused
  lattice point whose addition satisfies \Call{Legal}{}
  \For{each decoded seed state $S$}
    \While{$|S|<N$ and untested lattice points remain}
      \State traverse the remaining points in a fresh random order
      \State append the first $p$ with \Call{Legal}{$(S\cup\{p\})$}; permanently remove
      every tested point, including $p$, from the remaining pool
      \State stop if no tested point is legal
    \EndWhile
  \EndFor
  \State in each batch of $B$ terminal states, retain the top $qB$ states by cardinality
  \State $\mathcal{D}_{i+1}\gets$ retained states; save their unaugmented and augmented forms
\EndFor
\end{algorithmic}
\end{algorithm}

\begin{table}[t]
\caption{PatternBoost hyperparameters.}
\label{tab:pb-hparams}
\centering
\footnotesize
\setlength{\tabcolsep}{3pt}
\renewcommand{\arraystretch}{1.04}
\begin{tabular}{@{}>{\raggedright\arraybackslash}p{0.53\linewidth}
>{\raggedright\arraybackslash}p{0.43\linewidth}@{}}
\toprule
Parameter & Value \\
\midrule
\multicolumn{2}{@{}l@{}}{\textbf{Transformer architecture}}\\[-2pt]
Causal blocks / attention heads & $4$ / $4$ \\
Embedding / FFN width & $128$ / $512$ \\
Activation / normalization & GELU / pre-LN \\
Dropout & $0$ \\
\addlinespace
\multicolumn{2}{@{}l@{}}{\textbf{Training}}\\[-2pt]
Training rounds & $20$ \\
Retained configurations / round & $50$ \\
Optimizer & AdamW \\
Learning rate / schedule & $5\times10^{-4}$ / fixed \\
Weight decay & $0.01$ \\
Adam $\beta_1,\beta_2$ / $\epsilon$ & $(0.9,0.99)$ / $10^{-8}$ \\
Maximum updates / round & $500$ \\
Minibatch size & $32$ \\
Validation-orbit fraction & $0.1$ \\
\addlinespace
\multicolumn{2}{@{}l@{}}{\textbf{Proposal and saturation}}\\[-2pt]
Generated states / round & $250$ \\
Selection batch / keep fraction & $250$ / $0.2$ \\
Saturation workers & $16$ \\
Sampling temperature / top-$k$ & $1$ / none \\
\bottomrule
\end{tabular}
\end{table}

Teacher forcing pads every sequence to context length $N+1$ (12 in 3D and 22 in 4D).
For an ordered point-token
sequence $(\tau_1,\ldots,\tau_k)$, the input is
$[0,\tau_1,\ldots,\tau_k]$ and the target is
$[\tau_1,\ldots,\tau_k,0]$; target positions after the first stop are set to $-1$ and
ignored by cross-entropy. Minibatches are sampled uniformly with replacement.

Let $m_\ell$ be the multiplicity of side length $\ell$ in $L$. The augmentation group has
\[
  |\mathcal H(L)|=2^d\prod_{\ell}m_\ell!
\]
box-preserving transformations: independent reflections and permutations among axes with
equal side lengths. All transforms are emitted without deduplicating coincident arrays.
For $L=(20,20,5)$ this gives $16$ transforms; for $L=(6,6,6,6)$ it gives
$2^4 4!=384$. With $50$ retained configurations, the nominal $45/5$ orbit split yields
$720/80$ augmented train/validation sequences in 3D and $17{,}280/1{,}920$ in 4D when all
orbits are distinct. Complete symmetry orbits are assigned wholly to one split before
augmentation. Validation evaluates at most ten batches of $100$. Training runs for $500$
updates and evaluates every $25$ updates; the minimum validation-loss checkpoint among
steps $25,50,\ldots,500$ is used both for generation and to initialize the next round.
Exact split sizes and selected steps are recorded per generation.

The Transformer uses learned token and absolute-position embeddings, untied output
weights, causal self-attention, and PyTorch's default parameter initialization. Its dropout rate is zero. At the start of a new round, model weights come from the preceding round's
selected checkpoint but AdamW is reinitialized; optimizer state is restored only when
resuming an interruption within the same round. AdamW uses the fixed learning rate in
\Cref{tab:pb-hparams}.

Generation always draws $N$ tokens after the initial \textsc{Start}, without constraining
the vocabulary, in minibatches of $10$. Decoding stops at the first sampled
\textsc{Stop}; before that stop, duplicate or illegal point tokens are removed from the
remaining-point pool and skipped while later tokens are still attempted. Saturation
repeatedly draws a fresh permutation of the remaining pool, accepts the first legal point,
and permanently removes all points tested in that pass. The $250$ terminal states form one
selection batch; the largest $20\%$, namely $50$ states, survive. Stable input order breaks
cardinality ties. Generation and saturation use $16$ CPU workers. A master seed of
$20260809$ initializes Python, NumPy, and PyTorch randomness; per-state seeds are assigned
before parallel dispatch.

\subsection{AlphaEvolve-style program-evolution experiment configuration}
\label{app:ae-config}

\paragraph{Dimension-general task interface.}
For every instance $(d,N)$, the immutable wrapper receives the same typed inputs: sixteen
cataloged $(N-1)\times d$ arrays encoding Radon-free configurations, the
dimension-independent geometry routines, target size $N$, output cap, and remaining oracle
allowance. Returned candidates must have shape $N\times d$. The mutable four-function
interface, validation order, fitness, oracle allowance, output cap, and MAP-Elites mechanism
are shared across dimensions. The dimension, target, reference catalog, feature-bin count
for \texttt{best\_size}, and number of evolution mutations are recorded
separately below. No lattice side lengths are passed to the evolved program or enforced by
the evaluator.

\paragraph{Dimension-specific inputs.}
The 3D run samples eight references from each of two catalogs of verified 10-point
configurations; the 4D run samples sixteen references from a catalog of verified 20-point
configurations. Both use selection seed $2026$, run for \SearchOEMutations mutations, and
accept arbitrary finite integer output coordinates. The completed runs take 4.88 and 47.58
hours in 3D and 4D, respectively.

Each call to \texttt{generate\_candidates} installs process-local wrappers around the
geometry functions.  Calls to \texttt{find\_rps} while the candidate program is active
increment a counter and are limited jointly to $\SearchOECalls$; after exhaustion, a further
call returns a nonempty sentinel rather than permitting a false Radon-free result.
Convexity calls are uncharged.  The internal allowance closes before the external
evaluator begins, so rechecking the returned batch neither consumes nor enlarges the
program's search budget.  Every program receives a newly initialized NumPy
\texttt{default\_rng(42)} and may return at most $\SearchOECandidatesPerMutation$ arrays.

\paragraph{Candidate validation and fitness.}
The evaluator imports the generated module and invokes
\texttt{generate\_candidates(N,32)}; the declared three-dimensional instance instantiates
this call with $N=11$. Exceptions receive score $0$ and an error flag.
Each returned object is converted to a NumPy array and rejected unless it is finite,
two-dimensional with $d$ columns, integer-valued, and row-distinct. More than $\SearchOECandidatesPerMutation$
objects are truncated; arrays with a row count other than $N$ receive candidate score
$0$. For a target-sized array $X=(x_1,\ldots,x_N)$, let
$P=(x_1,\ldots,x_{N-1})$ denote its ordered prefix. The exact candidate score is
\begin{equation}
\label{eq:ae-score}
s(X)=
\begin{cases}
1, & X\text{ is convex and }\nrp(X)=0,\\
0.50+0.49/(1+\nrp(X)), & X\text{ is convex and }\nrp(X)>0,\\
0.40+0.09((N-1)/N), & X\text{ is not convex, but }P\text{ is convex and }\nrp(P)=0,\\
0, & \text{otherwise.}
\end{cases}
\end{equation}
The program fitness \texttt{combined\_score} is $\max_X s(X)$ over the returned batch.
The evaluator records the best candidate and summary batch statistics; no metric combines
properties from different candidates.

 \begin{algorithm}[H]
\caption{Dimension-general AlphaEvolve-style program-evolution search.}
\label{alg:alphaevolve-radon}
\begin{algorithmic}[1]
\Require instance $(d,N)$, initial program $p_0$, fixed evaluator $E$, islands
$\mathcal{I}$, number of mutations $T$, target fitness $1$
\State evaluate $p_0$ with $E$; insert its source, metrics, and features into the database
\For{$t=1,\ldots,T$}
  \State choose the active island and sample a parent from its population/archive
  \State assemble system and user messages from the fixed task prompt, parent source and
  metrics, execution artifacts, lineage, and selected top/diverse programs
  \State request an exact \texttt{SEARCH/REPLACE} diff from \texttt{gpt-5.6-sol}
  \State apply valid edits only inside the mutable \texttt{EVOLVE-BLOCK}; reject a full rewrite
  \State execute the child with a fresh RNG seed $42$ and a shared allowance of $\SearchOECalls$
  internal \texttt{find\_rps} calls
  \State $E$ independently validates and scores at most $\SearchOECandidatesPerMutation$ returned candidates
  \State store source, parent, prompt, response, metrics, features, and candidate artifact
  \State update the island and global archive; migrate after every ten island generations
  \State checkpoint; \textbf{stop} if \texttt{combined\_score}$=1$
\EndFor
\end{algorithmic}
\end{algorithm}

\begin{table}[H]
\caption{AlphaEvolve-style program-evolution settings.}
\label{tab:ae-hparams}
\centering
\footnotesize
\setlength{\tabcolsep}{3pt}
\begin{tabular}{@{}>{\raggedright\arraybackslash}p{0.53\linewidth}
>{\raggedright\arraybackslash}p{0.43\linewidth}@{}}
\toprule
Parameter & Value \\
\midrule
Output cap & $\SearchOECandidatesPerMutation$ candidates \\
Reference seeds & 16 cataloged $(N-1)$-point records \\
Reference-selection seed / program RNG & $2026$ / NumPy PCG64 seed $42$ \\
Internal oracle allowance & $\SearchOECalls$ \texttt{find\_rps} calls per program \\
Convexity calls / evaluator replay & uncharged / outside internal allowance \\
Evolution mutations / target score & $\SearchOEMutations$ per dimension / $1$ \\
Population / archive / islands & $24$ / $24$ / $3$ \\
Migration interval & $10$ island generations \\
MAP-Elites features / bins & \texttt{best\_size}: 3D $12$, 4D $22$; \texttt{convex\_fraction}: $10$ \\
Evolution mode & exact diffs; full rewrites disabled \\
LLM / weight & \texttt{gpt-5.6-sol} / $1.0$ \\
Reasoning effort / output limit & medium / $8{,}000$ tokens \\
\bottomrule
\end{tabular}
\end{table}

\paragraph{Outcomes.}
Each AlphaEvolve-style program-evolution experiment completes \SearchOEMutations\ mutations.

Every child in both runs returns \SearchOECandidatesPerMutation\ convex target configurations, uses exactly
\SearchOECalls\ internal oracle calls, and has evaluator error flag zero. Each lineage therefore
returns $\SearchOEMutations\times\SearchOECandidatesPerMutation=\SearchOECandidates$ candidates. Neither returns
a Radon-free target. The best 3D program (first seen at mutation \SearchOEThreeFirstBest)
and best 4D program (first seen at mutation \SearchOEFourFirstBest) each attain one Radon
point and fitness $0.745$.

\subsection{AlphaZero-Radon experiment configuration}
\label{app:alphazero-config}
\label{app:az-config}

The reported runs use the Plan-D search and training implementation. The 3D instance has
$L=(20,20,5)$, $N=11$, and $2{,}000$ actions. The 4D instance has $L=(6,6,6,6)$,
$N=21$, and $1{,}296$ actions. Both complete 2,000 iterations, comprising 200,000
generated trajectories per dimension; all completed iterations enter the reported outcomes.
The implementation calls this data-generation phase ``self-play,'' but the adapted task is
single-agent and has no opponent.

\paragraph{States, actions, and terminal outcomes.}
A state is the unordered set of placed lattice points, serialized internally by
mixed-radix tokens
\[
 a(x)=\sum_{j=1}^{d}x_j\prod_{k=1}^{j-1}L_k ,
\]
so the first axis varies fastest. Every game starts from the fixed origin
$(0,\ldots,0)$ and the agent selects the remaining $N-1$ points. The origin is an
explicit finite-box anchor, not a without-loss-of-generality reduction; it contributes
neither a policy target nor an edge reward. At an expanded node, occupied actions are
zeroed before candidate ranking. Applying an action terminates the branch when the new set
is non-convex, has positive $\nrp$, or reaches $N$. Rewards follow \Cref{eq:reward} and
are attached to the edge entering the successor. The value head estimates the undiscounted
future reward-to-go.

\paragraph{Plan-D top-$K$ PUCT.}
Let $P_\theta(\cdot\mid s)$ be the full-vocabulary softmax policy. Every newly expanded
node zeros occupied actions, retains the $K=64$ largest remaining probabilities, and
renormalizes them. Writing the retained support as $C(s)$, the tree prior is
\[
 \widehat P(a\mid s)=
 \frac{P_\theta(a\mid s)\mathbf 1\{a\in C(s)\}}
 {\sum_{b\in C(s)}P_\theta(b\mid s)} .
\]
At the root only, self-play replaces this prior by
$0.75\widehat P+0.25\eta$, where
$\eta\sim\operatorname{Dir}(0.1\mathbf 1)$. There is no eager root probe: a transition is
checked only when tree selection reaches it. Once a child is found terminal, Plan D marks
that edge and excludes it from later tree selections. Terminal means any non-convex or
Radon-positive set as well as a target success, so a scored terminal receives one backup
before it is masked. Search does not stop early upon discovering a success.

\begin{algorithm}[H]
\caption{AlphaZero-Radon single-agent trajectory-generation and training loop.}
\label{alg:alphazero}
\begin{algorithmic}[1]
\Require lattice sides $L$, target $N$, iterations $I$, games $G$, reads $M$,
candidate count $K$, $c_{\mathrm{puct}}$, Dirichlet parameter $\alpha$
\State initialize policy--value parameters $\theta$, Adam state, and FIFO buffer $\mathcal B$
\For{$i=1,\ldots,I$}
  \State $\mathcal D_i\gets\emptyset$
  \For{$g=1,\ldots,G$}
    \State initialize $s_0=\{(0,\ldots,0)\}$ and empty state, policy, and reward lists
    \While{$s_t$ is not terminal}
      \State $\pi_t\gets\Call{PUCTSearch}{s_t,\theta,M,K,c_{\mathrm{puct}},\alpha}$
      \State store $(s_t,\pi_t)$; sample $a_t\sim\pi_t$; apply $a_t$
      \State append the realized edge reward and continue unless terminal
    \EndWhile
    \State compute $G_t=\sum_{j=t}^{T_{\rm end}-1}r_j$ backwards
    \State add all $(s_t,\pi_t,G_t,n_{\rm rp}^{\rm final})$ to $\mathcal D_i$
  \EndFor
  \State append $\mathcal D_i$ to $\mathcal B$; retain its latest $40$ iterations
  \For{$h=1,\ldots,10$}
    \State sample a set-length bucket proportional to its buffered count
    \State draw at most $128$ examples from that bucket without replacement
    \State update $\theta$ with Adam on \Cref{eq:az-loss}
  \EndFor
  \State every 10 iterations, save the model weights
\EndFor
\end{algorithmic}
\end{algorithm}

\begin{algorithm}[H]
\caption{Plan-D fixed-candidate PUCT search used by the reported runs.}
\label{alg:az-puct}
\begin{algorithmic}[1]
\Require state $s$, network $\theta$, reads $M$, candidate count $K$,
$c_{\mathrm{puct}}$, Dirichlet parameter $\alpha$
\Function{PUCTSearch}{$s,\theta,M,K,c_{\mathrm{puct}},\alpha$}
  \State create a fresh root; retain and renormalize its occupancy-legal policy top-$K$
  \State mix Dirichlet noise over those root candidates
  \For{$m=1,\ldots,M$}
    \State descend over unmarked retained edges by maximizing
    $Q(u,a)+c_{\mathrm{puct}}U(u,a)$
    \State evaluate the selected successor with \Call{StepOutcome}{}
    \If{terminal}
      \State mark its parent edge terminal; back up its edge reward with leaf value $0$
    \Else
      \State retain the policy top-$K$ candidate set at the leaf; set leaf value to $V_\theta$
      \State back up intervening edge rewards plus the leaf value
    \EndIf
  \EndFor
  \State \Return the normalized full-vocabulary root visit counts $\pi$
\EndFunction
\end{algorithmic}
\end{algorithm}

\begin{table}[t]
\caption{AlphaZero-Radon hyperparameters shared across dimensions.}
\label{tab:az-hparams}
\centering
\footnotesize
\setlength{\tabcolsep}{3pt}
\renewcommand{\arraystretch}{1.04}
\begin{tabular}{@{}>{\raggedright\arraybackslash}p{0.53\linewidth}
>{\raggedright\arraybackslash}p{0.43\linewidth}@{}}
\toprule
Parameter & Value \\
\midrule
\multicolumn{2}{@{}l@{}}{\textbf{Search and trajectory generation}}\\[-2pt]
Self-play worker processes & $32$ \\
Training iterations & $2{,}000$ \\
Self-play games / iteration & $100$ \\
MCTS reads / move & $500$ \\
PUCT constant $c_{\mathrm{puct}}$ & $1.0$ \\
Candidate-set size $K$ & $64$ \\
Dirichlet $\alpha$ / noise fraction & $0.1$ / $0.25$ \\
Self-play temperature $T$ & $1$ \\
\addlinespace
\multicolumn{2}{@{}l@{}}{\textbf{Network architecture}}\\[-2pt]
Model width & $256$ \\
Nyquist-safe Fourier bands / axis & $3$ in both dimensions \\
SAB layers / attention heads & $4$ / $4$ \\
Attention activation / normalization & ReLU / LayerNorm \\
Policy/value hidden width & $256$ / $256$ \\
Head activation / dropout & GELU / $0$ \\
\addlinespace
\multicolumn{2}{@{}l@{}}{\textbf{Optimization}}\\[-2pt]
Optimizer & Adam \\
Learning rate / schedule & $10^{-4}$ / fixed \\
Weight decay & $10^{-4}$ \\
Adam $\beta_1,\beta_2$ / $\epsilon$ & $(0.9,0.999)$ / $10^{-8}$ \\
Maximum minibatch / updates per iteration & $128$ / $10$ \\
Replay-window length & $40$ iterations \\
Value-loss weight $\lambda_V$ & $20$ \\
Discount $\gamma$ & $1$ \\
\addlinespace
\multicolumn{2}{@{}l@{}}{\textbf{Reward shaping}}\\[-2pt]
Step / Radon / success constants & $1$ / $2$ / $25$ \\
Convexity-break constant & $0$ \\
\bottomrule
\end{tabular}
\end{table}

\paragraph{PUCT and training targets.}
PUCT uses
\[
 Q(s,a)=\frac{W(s,a)}{1+N(s,a)},\qquad
 U(s,a)=\widehat P(s,a)\frac{\sqrt{N(s)}}{1+N(s,a)} .
\]
The root visit count starts at zero. Each simulation accumulates
normalized per-edge rewards between the leaf and every ancestor and adds $V_\theta$ only at
a nonterminal leaf. After 500 simulations, the root visit counts are normalized over the
full vocabulary and the played action is sampled from them with $T=1$.

For a minibatch of size $B$, define the full-vocabulary policy by
\[
 p_\theta(a\mid s_b)=
 \frac{\exp \ell_\theta(s_b,a)}
 {\sum_{c\in\Lambda(L)}\exp \ell_\theta(s_b,c)}.
\]
The loss is
\begin{equation}
\label{eq:az-loss}
\mathcal L =
-\frac{1}{B}\sum_{b=1}^{B}\sum_{a\in\Lambda(L)}\pi_b(a)
\log p_\theta(a\mid s_b)
+\lambda_V\frac{1}{B}\sum_{b=1}^{B}
\bigl(V_\theta(s_b)-G_b\bigr)^2 .
\end{equation}
Here $\lambda_V=20$, $\pi_b$ is the normalized MCTS visit vector, and $G_b$ is the
undiscounted reward-to-go. Although $\pi_b$ is sparse because search expands top-$K$
actions, the softmax denominator is the full action vocabulary.

\paragraph{Network.}
Each coordinate is normalized as $u_j=x_j/(L_j-1)$ and mapped to the Fourier feature vector
\[
  [\sin(2^k\pi u_j),\cos(2^k\pi u_j)]_{k=0}^{B_d-1},
\]
where the Nyquist-safe $B_d=3$ in both runs, followed by a learned linear projection to
width $256$.
Four layer-normalized set-attention blocks with four
heads encode the placed-point set. Each multihead-attention block uses scaled dot-product
attention, a query residual, a ReLU output projection with a second residual, and
LayerNorm after each residual. Every candidate cell uses the same coordinate encoder to
form a cross-attention query over the set latents; a
$256\rightarrow256\rightarrow1$ GELU MLP produces its policy logit. The value head
mean-pools the set latents and applies an independent
$256\rightarrow256\rightarrow1$ GELU MLP, whose scalar output is used directly. All
linear layers use PyTorch defaults, and the dropout rate is zero.

\paragraph{Replay and optimization.}
One randomly initialized network and one Adam optimizer persist across all iterations.
At each iteration, every self-play game receives the latest evaluation-mode weights, and
every move constructs a fresh tree. Deterministic verifier outcomes may be memoized within a trajectory; this does not alter transitions, terminal labels, rewards, or the search budget.

For each decision after the origin, self-play stores the pre-move token set, the normalized
visit policy, the final Radon count, and the reward-to-go.

The learner appends the current iteration before training. For each of ten updates, it selects a set-length bucket
with probability proportional to its buffered sample count and draws
$\min(128,\lvert\text{bucket}\rvert)$ examples uniformly without replacement.
Thus each iteration makes at most $1{,}280$ replay draws, independent of the current FIFO
population; samples may reappear in later updates. Adam uses the fixed learning rate
$10^{-4}$, PyTorch defaults for $(\beta_1,\beta_2,\epsilon)$, and coupled
$\ell_2$ weight decay $10^{-4}$.

Both jobs run uninterrupted from a clean initialization to iteration 2,000. The global
initialization seed is $42$; per-game and replay-sampling seeds are deterministic functions
of the iteration and game indices. End-to-end wall times are approximately $85.9$ hours in
3D and $561.4$ hours in 4D.

\begin{figure}[t]
  \centering
  \includegraphics[width=\textwidth]{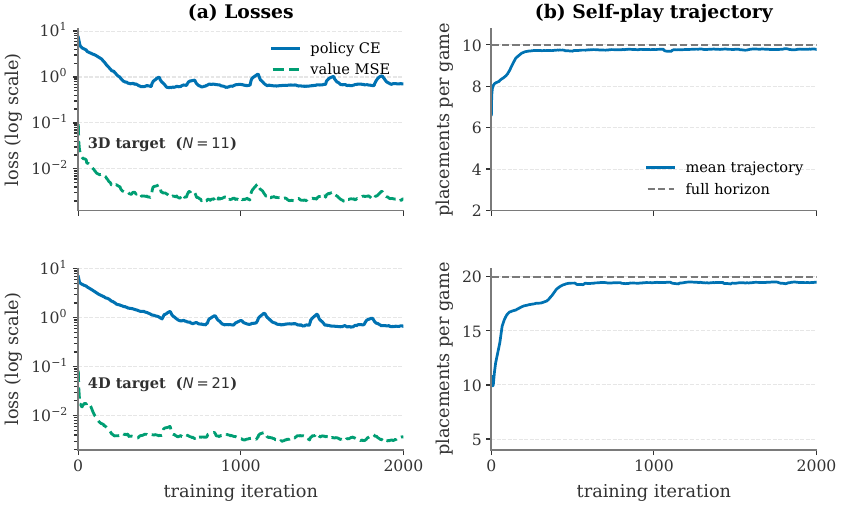}
  \caption{Training diagnostics for the complete 2,000-iteration AlphaZero-Radon
  runs. The top and bottom rows show the 3D ($N=11$) and 4D ($N=21$) instances,
  respectively. Curves are 50-iteration moving averages. The trajectory panels show
  only the mean trajectory length; the dashed line is the full horizon of 10 or 20
  policy-selected placements after the fixed origin.}
  \label{fig:az-training}
\end{figure}

\paragraph{Aggregation over the complete runs.}
In 3D, $157{,}485$ of 200,000 generated trajectories place an eleventh point. Of those terminals,
$8{,}651$ are convex and receive a complete Radon count; the minimum is $2$. In 4D,
$120{,}973$ games place a twenty-first point, $2{,}302$ of those terminals are convex, and
the minimum Radon count is $2$; a trajectory with terminal score two first appears at iteration
702. No target terminal has score zero. The remaining
target-cardinality terminals are non-convex and are not included in the ``target
candidates'' column. The largest fully verified prefixes have sizes 10 and 20,
respectively. \Cref{fig:az-training} reports the corresponding optimization and
trajectory-length diagnostics over the same complete runs.

\end{document}